\documentclass[letterpaper]{article} 
\usepackage{aaai24}  
\usepackage{times}  
\usepackage{helvet}  
\usepackage{courier}  
\usepackage[hyphens]{url}  
\usepackage{graphicx} 
\usepackage{natbib}  
\usepackage{caption} 
\usepackage{algorithm}
\usepackage{algorithmic}
\usepackage{multirow}
\usepackage{multicol}
\usepackage{graphicx}
\usepackage{amsmath}
\usepackage{amsfonts}
\usepackage{utfsym}
\usepackage{xcolor} 

\newcommand{\etal}{\textit{et al}. }
\newcommand{\ie}{\textit{i}.\textit{e}.}
\newcommand{\eg}{\textit{e}.\textit{g}.}

\usepackage{newfloat}
\usepackage{listings}
\DeclareCaptionStyle{ruled}{labelfont=normalfont,labelsep=colon,strut=off} 
\floatstyle{ruled}
\newfloat{listing}{tb}{lst}{}
\floatname{listing}{Listing}
\title{Learning Content-enhanced Mask Transformer for Domain Generalized Urban-Scene Segmentation}
\author{
    Qi Bi,
    Shaodi You,
    Theo Gevers
}
\affiliations{
    Computer Vision Research Group, University of Amsterdam, Netherlands \\

    \{q.bi, s.you, th.gevers\}@uva.nl
}

\usepackage{bibentry}

\begin{document}

\maketitle

\begin{abstract}
Domain-generalized urban-scene semantic segmentation (USSS) aims to learn generalized semantic predictions across diverse urban-scene styles. Unlike generic domain gap challenges, USSS is unique in that the semantic categories are often similar in different urban scenes, while the styles can vary significantly due to changes in urban landscapes, weather conditions, lighting, and other factors. 
Existing approaches typically rely on convolutional neural networks (CNNs) to learn the content of urban scenes.

In this paper, we propose a Content-enhanced Mask TransFormer (CMFormer) for domain-generalized USSS. The main idea is to enhance the focus of the fundamental component, the mask attention mechanism, in Transformer segmentation models on content information. 
We have observed through empirical analysis that a mask representation effectively captures pixel segments, albeit with reduced robustness to style variations. Conversely, its lower-resolution counterpart exhibits greater ability to accommodate style variations, while being less proficient in representing pixel segments. To harness the synergistic attributes of these two approaches, we introduce a novel content-enhanced mask attention mechanism. It learns mask queries from both the image feature and its down-sampled counterpart, aiming to simultaneously encapsulate the content and address stylistic variations. These features are fused into a Transformer decoder and integrated into a multi-resolution content-enhanced mask attention learning scheme.

Extensive experiments conducted on various domain-generalized urban-scene segmentation datasets demonstrate that the proposed CMFormer significantly outperforms existing CNN-based methods by up to 14.0\% mIoU and the contemporary HGFormer by up to 1.7\% mIoU. The source code is publicly available at \url{https://github.com/BiQiWHU/CMFormer}.
\end{abstract}

\section{Introduction}

\begin{figure}[!t]   
    \centering
    \includegraphics[width=1.0\linewidth]{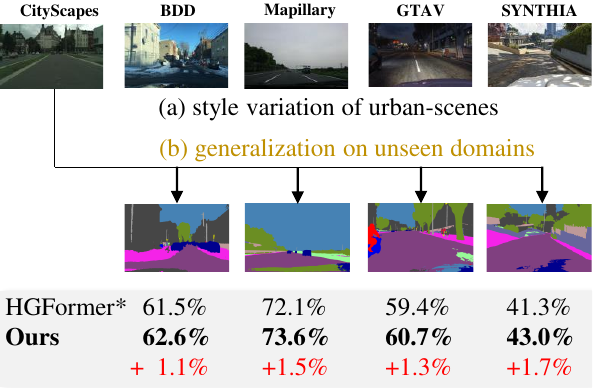}
\caption{(a) In domain generalized USSS, the domain gap is mainly from the extremely-varied styles. (b) A segmentation model is supposed to show good generalization on unseen target domains.}   
\label{motivation}   
\end{figure}

\begin{figure}[!t]   
    \centering
    \includegraphics[width=1.0\linewidth]{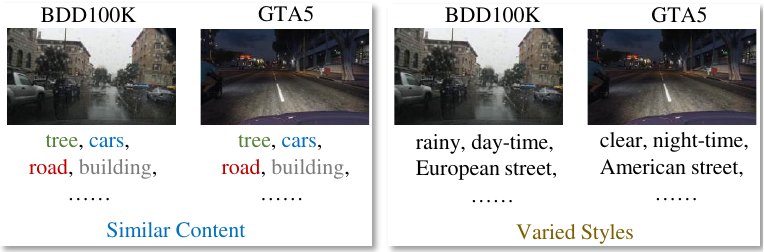}
\caption{Domain-generalized USSS demonstrates a distinctive feature of consistent content with diverse styles. An example is given for BDD100K and GTA5. }    
\label{CSobser}   
\end{figure}

Urban-scene semantic segmentation (USSS) is a challenging problem because of the large scene variations due to changing landscape, weather, and lighting conditions \cite{sakaridis2021acdc,mirza2022efficient,Bi2023interactive,chen2022learning}. 
Unreliable USSS can pose a significant risk to road users.
Nevertheless, a segmentation model trained on a specific dataset cannot encompass all urban scenes across the globe. As a result, the segmentation model is prone to encountering unfamiliar urban scenes during the inference stage.
Hence, domain generalization is essential for robust USSS
\cite{IBNet2018,instancenorm2019,Robust2021}, 
where a segmentation model can effectively extrapolate its performance to urban scenes that it hasn't encountered before (Fig.~\ref{motivation}).
In contrast to common domain generalization, domain generalized USSS requires special attention because the domain gap is mainly caused by large style variations whereas changes in semantics largely remain consistent (example in Fig.~\ref{CSobser}).

Existing approaches can be divided into two groups. One group focuses on the style de-coupling. This is usually achieved by a normalization \cite{IBNet2018,instancenorm2019,peng2022semantic} or whitening \cite{SW2019,Robust2021,xu2022dirl,peng2022semantic} transformation. However, the de-coupling methodology falls short as the content is not learnt in a robust way.
The other group is based on adverse domain training \cite{zhao2022style,lee2022wildnet,zhong2022adversarial}. However, these methods usually do not particularly focus on urban styles and therefore their performance is limited.

Recent work has shown that mask-level segmentation Transformer (e.g., Mask2Former) \cite{ding2023hgformer} is a scalable learner for domain generalized semantic segmentation. However, based on our empirical observations, a high-resolution mask-level representation excels at capturing content down to pixel semantics but is more susceptible to style variations.
Conversely, its down-sampled counterpart is less proficient in representing content down to pixel semantics but exhibits greater resilience to style variations.

A novel content-enhanced mask attention (CMA) mechanism is proposed. It jointly leverages both mask representation and its down-sampled counterpart, which show complementary properties on content representing and handling style variation.
Jointly using both features helps the style to be uniformly distributed while the content to be stabilized in a certain cluster. 
The proposed CMA takes the original image feature together with its down-sampled counterpart as input. Both features are fused to learn a more robust content from their complementary properties.

The proposed content-enhanced mask attention (CMA) mechanism can be integrated into existing mask-level segmentation Transformer in a learnable fashion.
It consists of three key steps, namely, exploiting high-resolution properties, exploiting low-resolution properties, and content-enhanced fusion.
Besides, it can also be seamlessly adapted to multi-resolution features.
A novel \textbf{C}ontent-enhanced \textbf{M}ask Trans\textbf{Former} (CMFormer) is proposed for domain-generalized USSS.

Large-scale experiments are
conducted with various domain generalized USSS settings, \ie, trained on one dataset from \cite{richter2016playing,ros2016synthia,cordts2016cityscapes,neuhold2017mapillary,yu2018bdd100k} as the source domain, and validated on the rest of the four datasets as the unseen target domains. 
All the datasets contain the same 19 semantic categories as the content, but vary in terms of scene styles.
The experiments show that the proposed CMFormer achieves up to 14.00\% mIoU improvement compared to the state-of-the-art CNN based methods (\eg, SAW \cite{peng2022semantic}, WildNet \cite{lee2022wildnet}).
Furthermore, it demonstrates a mIoU improvement of up to 1.7\% compared to the modern HGFormer model \cite{ding2023hgformer}.
It also shows state-of-the-art performance on synthetic-to-real and clear-to-adverse generalization.

Our contribution is summarized as follows:
\begin{itemize}
\item A content-enhanced mask attention (CMA) mechanism is proposed to leverage the complementary content and style properties from mask-level representation and its down-sampled counterpart.
\item On top of CMA, a Content-enhanced Mask Transformer (CMFormer) is proposed for domain generalized urban-scene semantic segmentation.
\item Extensive experiments show a large performance improvement over existing SOTA by up to 14.0\% mIoU, and HGFormer by up to 1.7\% mIoU.
\end{itemize}

\section{Related Work}

\noindent \textbf{Domain Generalization}
has been studied on no task-specific scenarios in the field of both machine learning and computer vision. 
Hu \etal \cite{hu2022feature} proposed a framework for image retrieval in an unsupervised setting. Zhou \etal \cite{zhou2020learning} proposed a framework to generalize to new homogeneous domains.
Qiao \etal \cite{qiao2020learning} and Peng \etal \cite{peng2022out} proposed to learn domain generalization from a single source domain. Many other methods have also been proposed \cite{zhao2020domain,mahajan2021domain,wang2020learning,chattopadhyay2020learning,segu2023batch}.

\noindent \textbf{Domain Generalized Semantic Segmentation} is more practical than conventional semantic segmentation \cite{pan2022label,ji2021learning,li2021joint,ji2021promoting,zhou2021differential,ye2021temporal}, which focuses on the generalization of a segmentation model on unseen target domains.
Existing methods focus on the generalization of in-the-wild \cite{piva2023empirical}, scribble \cite{tjio2022adversarial} and multi-source images \cite{kim2022pin,lambert2020mseg}, where substantial alterations can occur in both the content and style.

\noindent \textbf{Domain Generalized USSS}
focuses on the generalization of driving-scenes \cite{cordts2016cityscapes,yu2018bdd100k,neuhold2017mapillary,ros2016synthia,richter2016playing}. These methods use either normalization transformation (\eg, IBN \cite{IBNet2018}, IN \cite{instancenorm2019}, SAN \cite{peng2022semantic}) or whitening transformation (\eg, IW \cite{SW2019}, ISW \cite{Robust2021}, DIRL \cite{xu2022dirl}, SAW \cite{peng2022semantic}) on the training domain, to enable the model to generalize better on the target domains. Other advanced methods for domain generalization in segmentation typically rely on external images to incorporate more diverse styles \cite{lee2022wildnet,zhao2022style,zhong2022adversarial,li2023intra}, and leverage content consistency across multi-scale features \cite{PyramidConsistency2019}.
To the best of our knowledge, all of these methods are based on CNN.

\noindent \textbf{Mask Transformer for Semantic Segmentation}
uses the queries in the Transformer decoder to learn the masks, \eg,
Segmenter \cite{strudel2021segmenter}, MaskFormer \cite{cheng2021mask}. 
More recently, Mask2Former \cite{cheng2021per} further simplifies the pipeline of MaskFormer and achieves better performance.

\section{Preliminary}
\label{Sec3}
\paragraph{Problem Definition}
Domain generalization can be formulated as a worst-case problem \cite{li2021evaluating,zhong2022adversarial,volpi2018generalizing}. Given a source domain $\mathcal{S}$, and a set of unseen target domains $\mathcal{T}_1, \mathcal{T}_2, \cdots $, a model parameterized by $\theta$ with the task-specific loss $\mathcal{L}_{task}$, the generic domain generalization task can be formulated as a worst-case problem, given by
\begin{equation} \label{DGdef}
\setlength{\abovedisplayskip}{1pt}
\mathop{\rm{min}}\limits_{\theta} \mathop{\rm{supp}}\limits_{\mathcal{T}: D(\mathcal{S};\mathcal{T}_1, \mathcal{T}_2, \cdots) \leq \rho} \mathbb{E}_T [\mathcal{L}_{task}(\theta; \mathcal{T}_1, \mathcal{T}_2, \cdots)],
\setlength{\belowdisplayskip}{1pt}
\end{equation}
where $\theta$ denotes the model parameters, $D(\mathcal{S}; \mathcal{T}_1, \mathcal{T}_2, \cdots)$ corresponds to the distance between the source $\mathcal{S}$ and target domain $\mathcal{T}$, and $\rho$ denotes the constraint threshold. 

As summarized in ISW \cite{Robust2021}, domain generalized USSS is challenging as it has similar content (\eg, semantics) but may vary in
style (\eg, urban landscapes, weather, season, illumination) among each domain.



\begin{figure}[!t]
\centering
\includegraphics[width=0.47\textwidth]{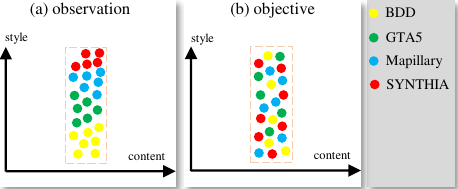} 
\caption{
(a) In the domain-generalized USSS setting, within the content-style space, samples from various domains tend to cluster closely along the content dimension while displaying dispersion along the style dimension. (b) An optimal generalized semantic segmentation scenario would involve uniform distribution of styles while maintaining content stability (as indicated by the brown bounding box).} 
\label{tsne}
\end{figure}

\paragraph{Content-style Feature Space}
Here we analyze the feature space.
Figure~\ref{tsne}a illustrates that in the context of domain-generalized USSS, samples from distinct domains might exhibit analogous patterns and cluster tightly along the content dimension. Conversely, samples from diverse domains may segregate into separate clusters along the style dimension.

An optimal and adaptable segmentation representation should achieve content stability while simultaneously exhibiting resilience in the face of significant style variations. Illustrated in Figure~\ref{tsne}b, our objective is to cultivate a content-style space wherein: 1) samples from diverse domains can occupy analogous positions along the content dimension, and 2) samples can be uniformly dispersed across the style dimension.
Both learning objectives allow us to therefore minimize the domain gap.

\paragraph{Overall Idea}

Recent work has shown that mask-level segmentation Transformer (e.g., Mask2Former) \cite{ding2023hgformer} is a scalable learner for domain generalized semantic segmentation. However, we empirically observe that, 
a mask-level representation is better at representing content, but more sensitive to style variations (similar to Fig.~\ref{tsne}a);
its low-resolution counterpart, on the contrary, is less capable to represent content, but more robust to the style variations (similar to the style dimensions in Fig.~\ref{tsne}b).

Overall, the mask representation and its down-sampled counterpart shows complementary properties when handling samples from different domains.
Thus, it is natural to jointly leverage both mask representation and its down-sampled counterparts, so as to at the same time stabilize the content and be insensitive to the style variation.

\paragraph{Difference between Existing Pipelines}

Existing methods usually focus on decoupling the styles from urban scenes, so that along the style dimension the samples from different domains are more uniformly distributed.

In contrast, the proposed method intends to leverage the content representation ability of mask-level features and the style handling ability of its down-sampled counterpart, so as to realize the aforementioned learning objective.

\section{Methodology}
\label{Sec4}
  
\subsection{Recap on Mask Attention}
\label{def3.1}

Recent studies show that the mask-level pipelines \cite{strudel2021segmenter,cheng2021mask,cheng2021per} have stronger representation ability than conventional pixel-wise pipelines for semantic segmentation, which can be attributed to the mask attention mechanism. 

It learns the query features as the segmentation masks by introducing a mask attention matrix based on the self-attention mechanism. Let $\mathbf{F}_l$ and $\mathbf{X}_l$ denote the image features from the image decoder and the features of the $l^{th}$ layer in a Transformer decoder, respectively.
When $l=0$, $\mathbf{X}_0$ refers to the input query features of the Transformer decoder. 

The key $\mathbf{K}_l$ and value $\mathbf{V}_l$ on $\mathbf{F}_{l-1}$ are computed by linear transformations $f_K$ and $f_V$, respectively.  
Similarly, the query $\mathbf{Q}_l$ on $\mathbf{X}_{l-1}$ is computed by linear transformation $f_Q$. 
Then, the query feature $\mathbf{X}_l$ is computed by
\begin{equation} 
\label{SelfAtt}
\mathbf{X}_l = {\rm softmax} (\mathcal{M}_{l-1} + \mathbf{Q}_l \mathbf{K}_{l}^\mathsf{T} ) \mathbf{V}_l + \mathbf{X}_{l-1},
\end{equation}
where $\mathcal{M}_{l-1} \in \{0,1\}^{N \times H_l W_l}$ is a binary mask attention matrix from the resized mask prediction of the previous $(l-1)^{th}$ layer, with a threshold of 0.5. $\mathcal{M}_{0}$ is binarized and resized from $\mathbf{X}_0$. It filters the foreground regions of an image, given by
\begin{equation} 
\label{attentionmask}
\mathcal{M}_{l-1}(x,y) =
\begin{cases}
0 & \text{if $\mathcal{M}_{l-1}(x,y)$=1 } \\
$ - \( \infty \) $ & \text{else}
\end{cases}.
\end{equation} 

\subsection{Exploiting High-Resolution Properties}
\label{sec4.1}

\begin{figure*}[!t]
  \centering
   \includegraphics[width=1.0\linewidth]{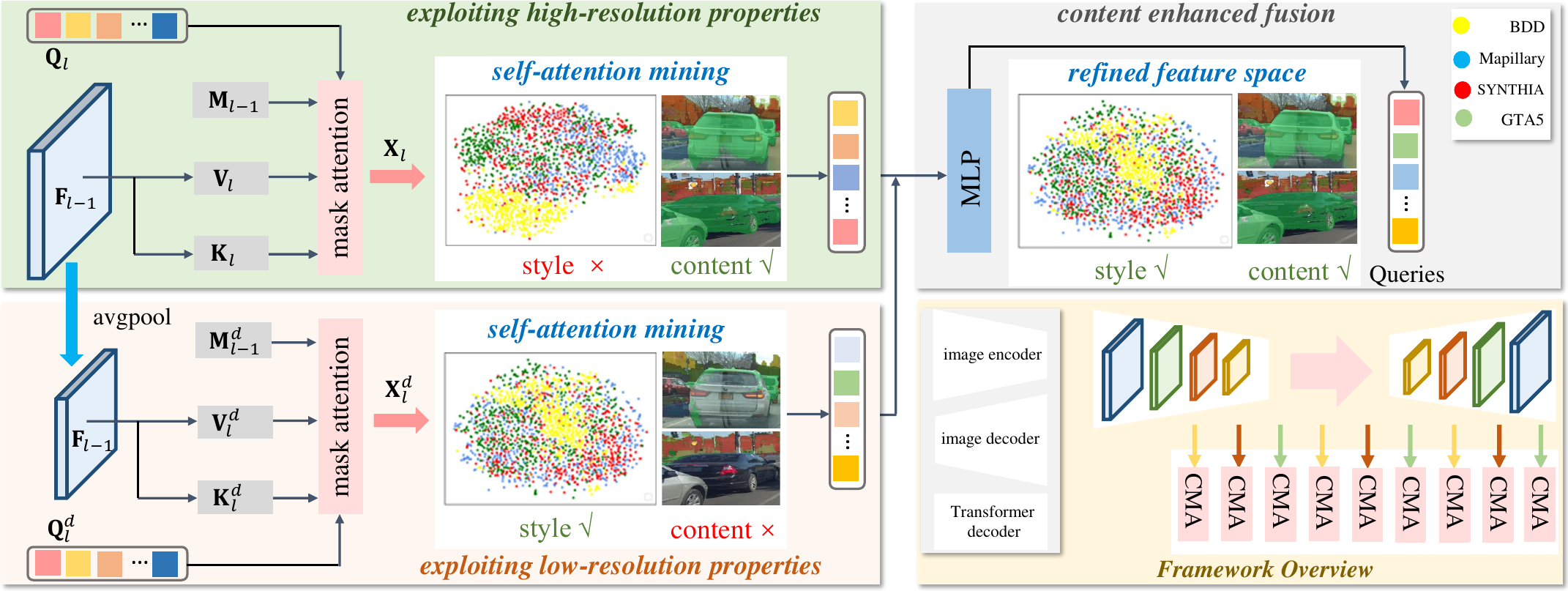}
   \vspace{-0.6cm}
   \caption{(a) The proposed Content-enhanced Mask Attention (CMA) consists of three key steps, namely, exploiting high-resolution properties (in green), exploiting low-resolution properties (in brown), and content enhanced fusion (in gray). (b) Framework overview (in yellow) of the proposed \textbf{C}ontent-enhanced \textbf{M}ask Trans\textbf{Former} (CMFormer) for domain generalized semantic segmentation. The image decoder is directly inherited from the Mask2Former \cite{cheng2021per}.
   }
   \label{framework}
\end{figure*}

Highlighted within the green block in Figure~\ref{framework}, our empirical observations reveal that the high-resolution mask representation exhibits the following characteristics: 1) greater proficiency in content representation, and 2) reduced susceptibility to domain variation. Achieving uniform mixing of samples from four domains presents a challenge.

To leverage the properties from high-resolution mask representations, we use the self-attention mechanism to exploit the amplified content representation from $\mathbf{X}_l$. Let $\mathbf{Q}_{\mathbf{X}_l}$, $\mathbf{V}_{\mathbf{X}_l}$ and $\mathbf{K}_{\mathbf{X}_l}$ denote its query, value and key, and $d_k$ denotes their dimension. Then, the self-attention is computed as
\begin{equation} \label{SA}
\footnotesize
{\rm Attention}(\mathbf{Q}_{\mathbf{X}_l}, \mathbf{K}_{\mathbf{X}_l}, \mathbf{V}_{\mathbf{X}_l}) = {\rm Softmax}(\frac{\mathbf{Q}_{\mathbf{X}_l} \mathbf{K}_{\mathbf{X}_l}}{\sqrt{d_k}})\mathbf{V}_{\mathbf{X}_l}, 
\end{equation}
where $\rm{Softmax}$ denotes the softmax normalization function, and the final output is denoted as $\tilde{\mathbf{X}}_l$. 

\subsection{Exploiting Low-Resolution Properties}

As shown in the brown block of Fig.~\ref{framework}, the low-resolution mask-level representation has the following properties: 
1) less qualified to represent the content; 
2) more capable to handle the style variation. 
In the feature space, samples from different domains are more uniformly distributed.
We propose to build a low-resolution mask representation derived from its high-resolution counterpart. This approach capitalizes on the attributes of the low-resolution representation to effectively address domain variations.

The low-resolution counterpart $\mathbf{F}_{l}^{d}$ is computed by average pooling $\rm{avgpool}$ from the original image feature $\mathbf{F}_{l}$ by
\begin{equation} 
\label{contentimagefeature}
\mathbf{F}_{l}^{d} = {\rm avgpool}(\mathbf{F}_{l}),
\end{equation} 
where the width and height of $\mathbf{F}_{l}$ is both twice the width and height of $\mathbf{F}_{l}^{d}$.

Similarly, the key and value from $\mathbf{F}_{l}^{d}$ are computed by linear transformations, and can be denoted as $\mathbf{K}^{d}_l$ and $\mathbf{V}^{d}_l$, respectively.
The query from $\mathbf{X}_{l-1}^{d}$ is also computed by linear transformation, and can be denoted as $\mathbf{K}^{d}_l$.
The mask attention on the low-resolution feature $\mathbf{X}_l^{d}$ is computed as
\begin{equation} 
\label{SelfAttContent}
\mathbf{X}_l^{d} = {\rm softmax} (\mathcal{M}_{l-1}^{d} + \mathbf{Q}_l \mathbf{K}_{l}^{d \mathsf{T}} ) \mathbf{V}_l^{d} + \mathbf{X}_{l-1}^{d}.
\end{equation}

To exploit the properties from the low-resolution mask representation $\mathbf{X}_l^d$, we use the self-attention mechanism. 
Let $\mathbf{Q}_{\mathbf{X}_l^d}$, $\mathbf{V}_{\mathbf{X}_l^d}$ and $\mathbf{K}_{\mathbf{X}_l^d}$ denote its query, value and keys. Then, the self-attention is computed by
\begin{equation} \label{SA}
\scriptsize
{\rm Attention}(\mathbf{Q}_{\mathbf{X}_l^d}, \mathbf{K}_{\mathbf{X}_l^d}, \mathbf{V}_{\mathbf{X}_l^d}) = {\rm Softmax}(\frac{\mathbf{Q}_{\mathbf{X}_l^d} \mathbf{K}_{\mathbf{X}_l^d}}{\sqrt{d_k}})\mathbf{V}_{\mathbf{X}_l^d}.
\end{equation}

The final output is denoted as $\tilde{\mathbf{X}}_l^d$.
It inherits the characteristics of the low-resolution mask representation, which is adept at accommodating style variations while being less resilient in capturing pixel-level intricacies.

\subsection{Content-enhanced Fusion}

Our idea is to leverage the complementary properties of mask-level representation and its down-sample counterpart, so as to enhance both the pixel-wise representing and style variation handing (shown in the gray box of Fig.~\ref{framework}).
The joint use of both representations
aids the segmentation masks in concentrating on scene content while reducing sensitivity to style variations.

To this end, we fuse both representations $\tilde{\mathbf{X}}_l$ and $\tilde{\mathbf{X}}_l^{d}$ in a simple and straight-forward way.
The fused feature $\mathbf{X}_l^{final}$ serves as the final output of the $l^{th}$ Transformer decoder, and it is computed as
\begin{equation}
\label{fusecontent}
\mathbf{X}_l^{final} = h_l([\tilde{\mathbf{X}}_l, \tilde{\mathbf{X}}_l^{d}]),
\end{equation}
where $[\cdot, \cdot]$ represents the concatenation operation, and $h_l(\cdot)$ refers to a linear layer.


\subsection{Network Architecture and Implementation Details}
\label{sec4.4}

The overall framework is shown in the yellow box of Fig.~\ref{framework}.
The Swin Transformer \cite{liu2021swin} is used as the backbone. 
The pre-trained backbone from ImageNet \cite{deng2009imagenet} is utilized for initialization. 

The image decoder from \cite{cheng2021per} uses the off-the-shelf multi-scale deformable attention Transformer (MSDeformAttn) \cite{zhu2021deformable} with the default setting in \cite{zhu2021deformable,cheng2021per}. 
By considering the image features from the Swin-Based encoder as input, every 6 MSDeformAttn layers are used to progressively up-sample the image features in $\times 32$, $\times 16$, $\times 8$, and $\times 4$, respectively. 
The $1/4$ resolution feature map is fused with the features from the Transformer decoder for dense prediction.

The Transformer decoder is also directly inherited from Mask2Former \cite{cheng2021per}, which has 9 self-attention layers in the Transformer decoder to handle the $\times 32$, $\times 16$ and $\times 8$ image features, respectively.

Following the default setting of MaskFormer \cite{cheng2021mask} and Mask2Former \cite{cheng2021per}, the final loss function $\mathcal{L}$ is a linear combination of the binary cross-entropy loss $\mathcal{L}_{ce}$, dice loss $\mathcal{L}_{dice}$, and the classification loss $\mathcal{L}_{cls}$, given by
\begin{equation}
\label{loss}
\mathcal{L} = \lambda_{ce} \mathcal{L}_{ce} + \lambda_{dice} \mathcal{L}_{dice} + \lambda_{cls} \mathcal{L}_{cls},
\end{equation}
with hyper-parameters $\lambda_{ce}=\lambda_{dice}=5.0, \lambda_{cls}=2.0$ as the default of Mask2Former without any tuning.
The Adam optimizer is used with an initial learning rate of $1\times10^{-4}$. The weight decay is set 0.05. 
The training terminates after 50 epochs. 

\section{Experiment}
\subsection{Dataset \& Evaluation Protocols}

Building upon prior research in domain-generalized USSS, our experiments utilize five different semantic segmentation datasets. Specifically, CityScapes \cite{cordts2016cityscapes} provides 2,975 and 500 well-annotated samples for training and validation, respectively. These driving-scenes are captured in Germany cities with a resolution of 2048$\times$1024.
BDD-100K \cite{yu2018bdd100k} also provides diverse urban driving scenes with a resolution of 1280$\times$720. 7,000 and 1,000 fine-annotated samples are provided for training and validation of semantic segmentation, respectively. 
Mapillary \cite{neuhold2017mapillary} is also a real-scene large-scale semantic segmentation dataset with 25,000 samples.
SYNTHIA \cite{ros2016synthia} is large-scale synthetic dataset, and provides 9,400 images with a resolution of 1280$\times$760.
GTA5 \cite{richter2016playing} is a synthetic semantic segmentation dataset rendered by the GTAV game engine. It provides 24,966 simulated urban-street samples with a resolution of 1914$\times$1052. We use C, B, M, S and G to denote these five datasets.

Following prior domain generalized USSS works \cite{IBNet2018,SW2019,Robust2021,peng2022semantic}, the segmentation model is trained on one dataset as the source domain, and is validated on the rest of the four datasets as the target domains. Three settings include: 1) G to C, B, M, S; 2) S to C, B, M, G; and 3) C to B, M, G, S. 
mIoU (in percentage \%) is used as the validation metric.
All of our experiments are performed three times and averaged for fair comparison. All the reported performance is directly cited from prior works under the ResNet-50 backbone \cite{IBNet2018,SW2019,Robust2021,peng2022semantic}.

Existing domain generalized USSS methods are included for comparison, namely, IBN \cite{IBNet2018}, IW \cite{SW2019}, Iternorm \cite{huang2019iterative}, DRPC \cite{PyramidConsistency2019}, ISW \cite{Robust2021}, GTR \cite{peng2021global}, DIRL \cite{xu2022dirl},  SHADE \cite{zhao2022style}, SAW \cite{peng2022semantic}, WildNet \cite{lee2022wildnet}, AdvStyle \cite{zhong2022adversarial}, SPC \cite{huang2023style}, and HGFormer \cite{ding2023hgformer}. 

\subsection{Comparison with State-of-the-art}

\renewcommand\arraystretch{1}
\begin{table}[!t]
	\centering
	\resizebox{\linewidth}{!}{
	\begin{tabular}{c|cccc}
		\hline
		\multirow{2}{*}{Method}  & \multicolumn{4}{c}{Trained on GTA5 (G)} \\ 
        \cline{2-5}
        ~ & $\rightarrow$ C & $\rightarrow$ B & $\rightarrow$ M & $\rightarrow$ S \\
        \hline
        IBN \cite{IBNet2018} & 33.85 & 32.30 & 37.75 & 27.90  \\
        IW  \cite{SW2019} & 29.91 & 27.48 & 29.71 & 27.61  \\
        Iternorm \cite{huang2019iterative} & 31.81 & 32.70 & 33.88 & 27.07 \\
        DRPC \cite{PyramidConsistency2019} & 37.42 & 32.14 & 34.12 & 28.06 \\
        ISW \cite{Robust2021} & 36.58 & 35.20 & 40.33 & 28.30  \\
        GTR \cite{peng2021global} & 37.53 & 33.75 & 34.52 & 28.17 \\
        DIRL \cite{xu2022dirl} & 41.04 & 39.15 & 41.60 & - \\ 
        SHADE \cite{zhao2022style} & 44.65 & 39.28 & 43.34 & - \\
        SAW \cite{peng2022semantic} & 39.75 & 37.34 & 41.86 & 30.79 \\
        WildNet \cite{lee2022wildnet} & 44.62 & 38.42 & 46.09 & 31.34 \\
        AdvStyle \cite{zhong2022adversarial} & 39.62 & 35.54 & 37.00 & - \\
        SPC  \cite{huang2023style} & 44.10 & 40.46 & 45.51 & - \\
        HGFormer \cite{ding2023hgformer} & - & - & - & - \\
        \hline
		\textbf{CMFormer} (Ours) & \textbf{55.31} & \textbf{49.91} & \textbf{60.09} & \textbf{43.80} \\ 
        ~ & \textcolor{red}{$\uparrow$10.66} & \textcolor{red}{$\uparrow$9.45} & \textcolor{red}{$\uparrow$14.00} & \textcolor{red}{$\uparrow$12.46} \\
        \hline
	\end{tabular}
        }
        \vspace{-0.2cm}
        \caption{G  $\rightarrow$ \{C, B, M, S\} setting. Performance comparison of the proposed CMFormer compared to existing domain generalized USSS methods. The symbol '-' indicates cases where the metric is either not reported or the official source code is not available by the submission due. Evaluation metric mIoU is given in ($\%$).}
	\label{DGsegGTA}
\end{table}

\textbf{GTA5 Source Domain}
Table~\ref{DGsegGTA}~reports the performance on target domains of C, B, M and S, respectively. 
The proposed CMFormer shows a performance improvement of 10.66\%, 9.45\%, 14.00\% and 12.46\% compared to existing state-of-the-art CNN based methods on each target domain, respectively.
These outcomes demonstrate the feature generalization ability of the proposed CMFormer. 
Notice that the source domain GTA5 is a synthetic dataset, while the target domains are real images. 
It further validates the performance of the proposed method.

\textbf{SYNTHIA Source Domain}
Table~\ref{DGsegsyn}~reports the performance. 
The proposed CMFormer shows a 5.67\%, 8.73\% and 11.49\% mIoU performance gain against the best CNN based methods, respectively.
However, on the BBD-100K (B) dataset, the semantic-aware whitening (SAW) method \cite{peng2022semantic} outperforms the proposed CMFormer by 1.80\% mIoU. 
Nevertheless, the proposed CMFormer still outperforms the rest methods. 
The performance gain of the proposed CMFormer when trained on SYNTHIA dataset is not as significant as it is trained on CityScapes or GTA5 dataset. 
The explanation may be that the SYNTHIA dataset has much fewer samples than GTA5 dataset, \ie, 9400 v.s. 24966, and a transformer may be under-trained. 

\renewcommand\arraystretch{1}
\begin{table}[!t]
	\centering
	\resizebox{\linewidth}{!}{
	\begin{tabular}{c|cccc}
		\hline
		\multirow{2}{*}{Method}  & \multicolumn{4}{c}{Trained on SYNTHIA (S)} \\ 
        \cline{2-5}
        ~ & $\rightarrow$ C & $\rightarrow$ B & $\rightarrow$ M & $\rightarrow$ G \\
        \hline
        IBN \cite{IBNet2018} & 32.04 & 30.57 & 32.16 & 26.90  \\
        IW \cite{SW2019} & 28.16 & 27.12 & 26.31 & 26.51  \\
        Iternorm \cite{huang2019iterative} & - & - & - & - \\
        DRPC \cite{PyramidConsistency2019} & 35.65 & 31.53 & 32.74 & 28.75 \\
        ISW \cite{Robust2021} & 35.83 & 31.62 & 30.84 & 27.68  \\
        GTR \cite{peng2021global} & 36.84 & 32.02 & 32.89 & 28.02 \\
        DIRL \cite{xu2022dirl} & - & - & - & - \\ 
        SHADE \cite{zhao2022style} & - & - & - & - \\
        SAW \cite{peng2022semantic} & 38.92 & \textbf{35.24} & 34.52 & 29.16 \\
        WildNet \cite{lee2022wildnet} & - & - & - & - \\
        AdvStyle \cite{zhong2022adversarial} & 37.59 & 27.45 & 31.76 & - \\
        SPC \cite{huang2023style} & - & - & - & - \\
        HGFormer \cite{ding2023hgformer} & - & - & - & - \\
        \hline
		CMFormer (Ours) & \textbf{44.59} & 33.44 & \textbf{43.25} & \textbf{40.65} \\ 
        ~ & \textcolor{red}{$\uparrow$5.67} & \textcolor{blue}{$\downarrow$1.80} & \textcolor{red}{$\uparrow$8.73} & \textcolor{red}{$\uparrow$11.49} \\
        \hline
	\end{tabular}
        }
        \vspace{-0.2cm}
        \caption{S  $\rightarrow$ \{C, B, M, G\} setting. Performance comparison of the proposed CMFormer compared to existing domain generalized USSS methods. The symbol '-' indicates cases where the metric is either not reported or the official source code is not available by the submission due. Evaluation metric mIoU is given in ($\%$).}
	\label{DGsegsyn}
\end{table}

\textbf{CityScapes Source Domain}
Table~\ref{DGsegcity}~reports the performance. 
As HGFormer only reports one decimal results \cite{ding2023hgformer}, we also report one decimal results when compared with it.
The proposed CMFormer (with Swin-Base backbone) shows a performance gain of 6.32\%, 10.43\%, 9.50\% and 12.11\% mIoU on the B, M, G and S dataset against the state-of-the-art CNN based method.
As BDD100K dataset contains many nigh-time urban-street images, it is particularly challenging for existing domain generalized USSS methods. 
Still, a performance gain of 6.32\% is observed by the proposed CMFormer. 

On the other hand, when comparing ours with the contemporary HGFormer with the Swin-Large backbone, it shows an mIoU improvement of 1.1\%, 1.5\%, 1.3\% and 1.7\% on the B, M, G and S target domain, respectively.

\renewcommand\arraystretch{1}
\begin{table}[!t]
	\centering
	\resizebox{\linewidth}{!}{
	\begin{tabular}{c|c|cccc}
		\hline
		\multirow{2}{*}{Method}  &  \multirow{2}{*}{Backbone}  & \multicolumn{4}{c}{Trained on Cityscapes (C)} \\ 
        \cline{3-6}
        ~ & ~ & $\rightarrow$ B & $\rightarrow$ M & $\rightarrow$ G & $\rightarrow$ S \\
        \hline
        IBN  & ResNet50 & 48.56 & 57.04 & 45.06 & 26.14  \\
        IW & ResNet50 & 48.49 & 55.82 & 44.87 & 26.10  \\
        Iternorm & ResNet50 & 49.23 & 56.26 &  45.73 & 25.98 \\
        DRPC & ResNet50 & 49.86 & 56.34 & 45.62 & 26.58 \\
        ISW & ResNet50 & 50.73 & 58.64 & 45.00 & 26.20  \\
        GTR & ResNet50 & 50.75 & 57.16 & 45.79 & 26.47 \\
        DIRL & ResNet50 & 51.80 & - & 46.52 & 26.50 \\ 
        SHADE & ResNet50 & 50.95 & 60.67 & 48.61 & 27.62 \\
        SAW & ResNet50 & 52.95 & 59.81 & 47.28 & 28.32  \\
        WildNet & ResNet50 & 50.94 & 58.79 & 47.01 & 27.95\\
        AdvStyle & ResNet50 & - & - & - & - \\
        SPC & ResNet50 & - & - & - & - \\
        \hline
        Mask2Former$\dag$ &  Swin-T & 51.3 & 65.3 & 50.6 & 34.0 \\
        HGFormer$\dag$ & Swin-T & 53.4 & 66.9 & 51.3 & 33.6 \\
        \hline
        Mask2Former & Swin-B & 55.43 & 66.12 & 55.05 & 38.19 \\
		CMFormer (Ours) & Swin-B & \textbf{59.27} & \textbf{71.10} & \textbf{58.11} & \textbf{40.43} \\ 
        \hline
         Mask2Former$\dag$ & Swin-L & 60.1 & 72.2 & 57.8 & 42.4 \\
        HGFormer$\dag$ & Swin-L & 61.5 & 72.1 & 59.4 & 41.3 \\
		CMFormer (Ours) & Swin-L & \textbf{62.6} & \textbf{73.6} & \textbf{60.7} & \textbf{43.0} \\ 
        \hline
	\end{tabular}
        }
        \vspace{-0.2cm}
        \caption{C  $\rightarrow$ \{B, M, G, S\} setting.
        Performance comparison of the proposed CMFormer compared to existing domain generalized USSS methods. The symbol '-' indicates cases where the metric is either not reported or the official source code is not available by the submission due. Evaluation metric mIoU is given in ($\%$).
        $\dag$: HGFormer only reports one decimal results \cite{ding2023hgformer}.
        }
	\label{DGsegcity}
\end{table}

\textbf{From Synthetic Domain to Real Domain}
We also test the generalization ability of the CMFormer when trained on the synthetic domains (G+S) and validated on the three real-world domains B, C and M, respectively. The results are shown in Table~\ref{genGStoreal}.
The proposed CMFormer significantly outperforms the instance normalization based (IBN \cite{IBNet2018}), whitening transformation based (ISW \cite{Robust2021}) and adversarial domain training based (SHADE \cite{zhao2022style}, AdvStyle \cite{zhong2022adversarial}) methods by $\textgreater$10\% mIoU.

\renewcommand\arraystretch{1}
\begin{table}[!t]
	\centering
	\begin{tabular}{c|ccc|c}
		\hline
		\multirow{2}{*}{Backbone} & \multicolumn{4}{c}{Trained on Two Synthetic Domains (G+S)} \\ 
        \cline{2-5}
        ~ & $\rightarrow$ Citys & $\rightarrow$ BDD & $\rightarrow$ MAP & mean \\
        \hline
        ResNet-50 & 35.46 & 25.09 & 31.94 & 30.83 \\
        IBN & 35.55 & 32.18 & 38.09 & 35.27  \\
        ISW & 37.69 & 34.09 & 38.49 & 36.75\\
        SHADE & 47.43 & 40.30 & 47.60 & 45.11\\
        AdvStyle & 39.29 & 39.26 & 41.14 & 39.90 \\
        SPC & 46.36 & 43.18 & 48.23 & 45.92 \\
        \hline
		Ours & \textbf{59.70} & \textbf{53.36} & \textbf{61.61} & \textbf{58.22} \\ 
        \hline
	\end{tabular}
        \vspace{-0.2cm}
        \caption{Generalization of the proposed CMFormer when trained on two synthetic datasets and generalized on real domains. Evaluation metric mIoU is presented in ($\%$).}
	\label{genGStoreal}
\end{table}

\textbf{From Clear to Adverse Conditions}
we further validate the proposed CMFormer's performance on the adverse conditions dataset with correspondance (ACDC) \cite{sakaridis2021acdc}. 
We set the fog, night, rain and snow as four different unseen domains, and directly use the model pre-trained on CityScapes for inference.
The results are shown in Table~\ref{genacdc}. 
It significantly outperforms existing domain generalized segmentation methods \cite{IBNet2018,instancenorm2019,SW2019,Robust2021,li2023intra} by up to 10.3\%, 0.5\%, 11.6\%, 11.1\% on the fog, night, rain and snow domains, respectively.

\renewcommand\arraystretch{1}
\begin{table}[!t]
	\centering
	\resizebox{\linewidth}{!}{
	\begin{tabular}{c|c|cccc|c}
		\hline
		\multirow{2}{*}{Method} & \multirow{2}{*}{Backbone} & \multicolumn{5}{c}{Trained on Cityscapes (C)} \\ 
        \cline{3-7}
        ~ & ~ & $\rightarrow$ Fog & $\rightarrow$ Night & $\rightarrow$ Rain & $\rightarrow$ Snow & mean \\
        \hline
        IBN & ResNet-50 & 63.8 & 21.2 & 50.4 & 49.6 & 43.7  \\ 
        Iternorm & ResNet-50 & 63.3 & 23.8 & 50.1 & 49.9 & 45.3 \\
        IW & ResNet-50 & 62.4 & 21.8 & 52.4 & 47.6 & 46.6  \\
        ISW & ResNet-50 & 64.3 & 24.3 & 56.0 & 49.8 & 48.1  \\ 
        ISSA & MiT-B3 & 67.5 & 33.2 & 55.9 & 53.2 & 52.5\\
        \hline
		CMFormer (Ours) & Swin-B & \textbf{77.8} & \textbf{33.7} & \textbf{67.6} & \textbf{64.3} & \textbf{60.1}\\ 
        \hline
	\end{tabular}
        }
        \caption{Generalization of the proposed CMFormer to the adverse condition domains (rain, fog, night and snow) on ACDC dataset \cite{sakaridis2021acdc}. The mean mIoU (presented in $\%$) of all the four domains is NOT a simple average of mIoU.}
	\label{genacdc}
\end{table}

\begin{table}[!t]
\footnotesize
\begin{center}
\label{ablation}
\resizebox{\linewidth}{!}{
\begin{tabular}{ccc|cccc|cccc}
\hline
\multicolumn{3}{c|}{Content Enhancement} & \multicolumn{4}{c|}{Trained on CityScapes (C)} & \multicolumn{4}{c}{ 
Trained on SYNTHIA (S)} \\
\hline
$\times$32 & $\times$16 & $\times$8 & $\rightarrow$ B & $\rightarrow$ M & $\rightarrow$ G & $\rightarrow$ S & $\rightarrow$ C & $\rightarrow$ B & $\rightarrow$ M & $\rightarrow$ G \\
\hline
~ & ~ & ~ & 55.43 & 66.12 & 55.05 & 38.19 & 42.64 & 31.91 & 36.80 & 37.29 \\
$\checkmark$ & ~ & ~ & 56.17 & 67.55 & 55.42 & 38.83 & 42.77 & 31.62 & 41.61 & 37.56 \\
$\checkmark$ & $\checkmark$ & ~ & 58.10 & 69.72 & 55.54 & 39.41 & 44.22 & 33.19 & 42.23 & 38.78 \\
$\checkmark$ & $\checkmark$ & $\checkmark$ & \textbf{59.27} & \textbf{71.10} & \textbf{58.11} & \textbf{40.43} & \textbf{44.59} & \textbf{33.44} & \textbf{43.25} & \textbf{40.65} \\
\hline
\end{tabular}
}
\caption{Ablation studies on each component of the proposed CMFormer. $\times$32, $\times$16 and $\times$8 denote the image features of $\times$32, $\times$16 and $\times$8 resolution. $\checkmark$ refers to the content enhancement is implemented. Evaluation metric mIoU.}
\label{ablation}
\end{center} 
\end{table}

\begin{figure*}[!t]
  \centering
   \includegraphics[width=1.0\linewidth]{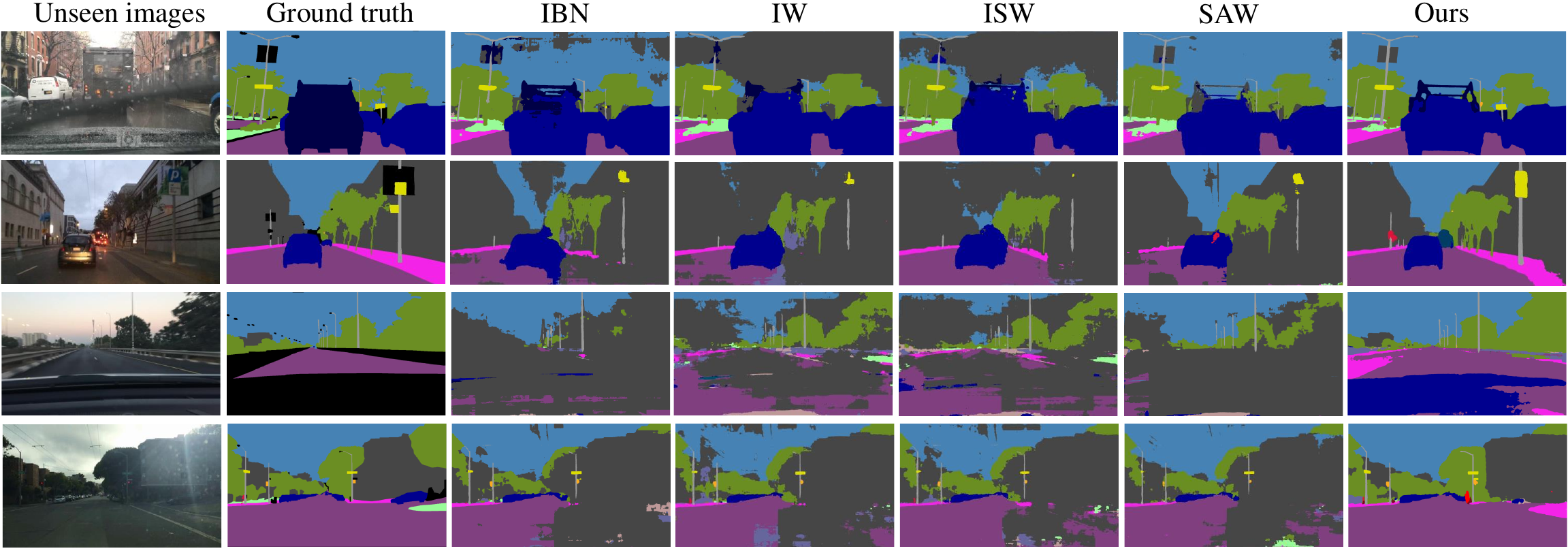}
   \vspace{-0.7cm}
   \caption{Unseen domain segmentation prediction of existing CNN based domain generalized semantic segmentation methods (IBN \cite{IBNet2018}, IW \cite{SW2019}, ISW \cite{Robust2021}, SAW \cite{peng2022semantic}) and the proposed CMFormer under the C $\rightarrow$ B, M, G, S setting. 
   }
   \label{visinfer}
\end{figure*}

\begin{figure*}[!t]
  \centering
   \includegraphics[width=1.0\linewidth]{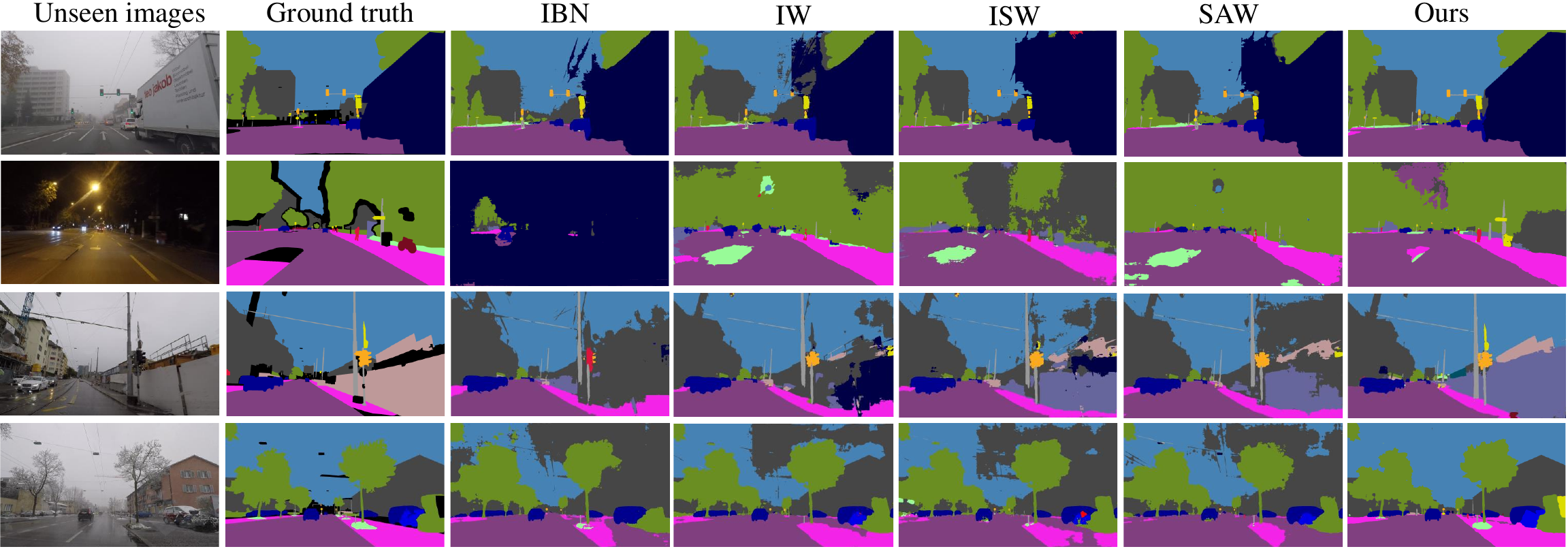}
   \vspace{-0.7cm}
   \caption{Unseen domain segmentation prediction of existing CNN based domain generalized semantic segmentation methods (IBN \cite{IBNet2018}, IW \cite{SW2019}, ISW \cite{Robust2021}, SAW \cite{peng2022semantic}) and the proposed CMFormer under the C $\rightarrow$ adverse domain setting. 
   }
   \label{visinfer2}
\end{figure*}

\subsection{Ablation Studies}

\textbf{On Content-enhancement of Each Resolution} Table~\ref{ablation}~reports the performance of the proposed CMFormer when $\times 32$, $\times 16$ and $\times 8$ image features are or are not implemented with content enhancement. 
The content enhancement on a certain resolution feature allows the exploiting of its low-resolution properties.
When no image features are implemented with content enhancement, CMFormer degrades into a Mask2Former \cite{cheng2021per} which only includes the high-resolution properties. 
When only implementing content enhancement on the $\times 32$ image feature, the down-sampled $\times 128$ image feature may propagate little content information to the segmentation mask, and only a performance gain of 0.74\%, 1.43\%, 0.37\% and 0.64\% on B, M, G and S target domain is observed.
When further implementing content enhancement on the $\times 16$ image feature, the enhanced content information begins to play a role, and an additional performance gain of 1.93\%, 2.17\%, 0.12\% and 0.58\% is observed. 
Then, the content enhancement on the $\times 8$ image feature also demonstrates a significant impact on the generalization ability. 
Similar observation can be found on the S$\rightarrow$C, B, M, G setting.

\subsection{Quantitative Segmentation Results}

Some segmentation results on the 
C $\rightarrow$ B, M, G, S setting and C $\rightarrow$ adverse domain setting are visualized in Fig.~\ref{visinfer}~and~\ref{visinfer2}. 
Compared with the CNN based methods, the proposed CMFormer shows a better segmentation prediction, especially in terms of the completeness of objects.

\section{Conclusion}
\label{Sec6}

In this paper, we explored the feasibility of adapting the mask Transformer for domain-generalized urban-scene semantic segmentation (USSS). To address the challenges of style variation and robust content representation, we proposed a content-enhanced mask attention (CMA) mechanism. 
This mechanism is designed to capture more resilient content features while being less sensitive to style variations. 
Furthermore, we extended it to incorporate multi-resolution features and integrate it into a novel framework called the \textbf{C}ontent-enhanced \textbf{M}ask Trans\textbf{Former} (CMFormer).
Extensive experiments on multiple settings demonstrated the superior performance of CMFormer compared to existing domain-generalized USSS methods. 


\bibliography{aaai24}

\end{document}